# Constraint Logic Programming for Natural Language Processing

Philippe Blache
2LC - CNRS

1361, route des Lucioles
F-06560 Sophia Antipolis
pb@llaor.unice.fr

Nabil Hathout
INaLF - CNRS
Château du Montet
Rue du Doyen Roubault
F-54500 Vandœuvre-lès-Nancy
hathout@ciril.fr

**Abstract**

This paper proposes an evaluation of the adequacy of the constraint logic programming paradigm for natural language processing. Theoretical aspects of this question have been discussed in several works (see for example [Smolka89] or [Aït-Kaci92]). We adopt here a pragmatic point of view and our argumentation relies on concrete solutions. Using actual contraints (in the CLP sense) is neither easy nor direct. However, CLP can improve parsing techniques in several aspects such as concision, control, efficiency or direct representation of linguistic formalism. This discussion is illustrated by several examples and the presentation of an HPSG parser.

## 1 Introduction

Contemporary linguistic theories generally describe syntactic properties as constraints on linguistic structures. This constraint-based paradigm (see [Shieber92]) is used both to define the set of possible structures (trees, feature structures, etc.) and to reduce the domain to the set of well-formed structures.

From a computational point of view, there is a clear distinction between active and passive constraints. Usually, logic programming employs a *generate-and-test* technique in which variable values are generated before their properties are verified. Search space is reduced *a posteriori* according to what is termed *passive handling* of constraints. On the other hand, constraint logic programming is based on *a priori* reduction of the search space upon so called *active constraints* (see [Cohen90] or [VanHentenryck89]). As concerns natural language processing, constraint logic programming (hereafter CLP) can improve classic implementations in two directions: efficiency (better control of parsing process and reduction of non-determinism) and knowledge representation (concision and direct implementation of linguistic constraints as computational constraints).

More precisely, CLP allows a direct interpretation of constraint-based theories by implementing *linguistic constraints* as *active constraints*. However the question is more complex than expected for actual linguistic formalisms. In particular, the main problem comes from the fact that the model (i.e. the syntactic structure) in which constraints must be interpreted is unknown before the resolution. In this case, constraints usually become passive and the resolution method turns out to be classic generate-and-test (see [Blache92], [Hathout94]). This is the solution implemented by most of the constraint-based NLP systems like, for example, ALE (cf. [Carpenter92], [Carpenter94]) or HPSG-PL (cf. [Popowitch91], [Kodrič92]). In these cases, even if linguistic principles are represented as constraints on the structures, they are implemented (compiled) as classic Prolog rules. In fact, the only kind of constraint used there is unification. We show that CLP offers several other constraints which can be very useful.

In this paper, we also discuss the nature of the dependency between active constraints and linguistic structures. Only constraints insensitive to model underspecification can be easily implemented as active constraints. Such an implementation is more complex for the others: their resolution requires the repetition of computations determining whether the usual CLP resolution of the constraint can start or not. Roughly speaking, the first type of constraints corresponds to principles involving local relations or relations between objects connected by "paths" while the later corresponds to long distance relations between objects identified by a set of properties. Roughly, we can distinguish two kinds of constraints according to their scope:

- syntactic structure constraints, and
- value specification constraints.

The former concern the syntactic structure itself whereas the latter specify constraints on the features of "atomic" objects such as tree nodes. From a practical point of view, the evaluation of the adequacy of CLP for natural language processing can be achieved in two ways: one concerning the implementation level and another the theoretical one. Each level induces different problems and relies on different CLP resources. Combining these resources can improve NLP systems in several ways: efficiency, coverage, control and adequacy with the theoretical framework.

The first section reviews some problems and describes the limits of the active handling of structural constraints. The second section presents a little context-free grammar parser with which we show how a parsing problem can be seen as a constraint satisfaction problem. The two last sections deal with the implementation of linguistic constraints.

## 2 Limits of the Active Handling of Structural Constraints

In this section, we are concerned with the main problems induced by the active handling of constraints which are sensitive to model underspecification. This class of constraints includes all *structural constraints*, such as, for instance, `has` proposed in [Saint-Dizier91]. `has(X,Y)` is a long distance constraint on feature structures `X` and `Y` which imposes to `Y` to be a substructure of `X` (somewhere inside it); in other words, `Y` must be the value of a path starting at `X`. In the general case, `has(X,Y)` cannot be solved before `X` becomes ground.

Structural constraints posited by a NLP system (e.g. a parser) on a linguistic structure are more complex to solve than constraints over standard CLP domains (boolean, numeral, finite domains) because they must be interpreted with respect to a model which is the very linguistic structure being built by the NLP system, and in general, this structure becomes *completely* known only when the current processing ends.

However, structural constraints should be taken into account by the solver as soon as they are posited by the NLP system, even if the model is still uncomplete.

## 2.1 Description of the Model

The linguistic structures built by an NLP change along with the input of the CLP program (e.g. the input sentence, in the case of a parser), at each new execution. Therefore, constraints over linguistic structures may be interpreted with respect to a set of different models. In consequence, CLP systems designed for handling and resolving these constraints are not instances of Jaffar and Lassez's *CLP schema* [Jaffar86]. On the other hand, their modelization can be formulated in the framework proposed in [Höhfeld88].

In order to be able to solve the constraints posited by the NLP system, the solver needs to know with respect to which is the model these constraints have to be interpreted. This model being unknown beforehand, the NLP system must describe it to the solver while it is building it. This requirement is indeed a drawback because it greatly reduces the concision of CLP programs since it increases the amount of information to be given to the solver. The CLP constraints language used for the implementation of the NLP system must allow the user both to posit constraints on variables and to describe the model to be used for their resolution. The distinction between these two kinds of information can be formalized by means of the *Ask & Tell* paradigm proposed by [Saraswat90]: the model is performed by means of *tell* constraints whereas structural constraints (to be solved) are expressed as *ask* constraints.

## 2.2 Structural Constraints Resolution

The resolution of a constraint system with respect to an underspecified model is done in two stages: first, selection of the subsystem of resolvable constraints, then standard CLP resolution of this subsystem. The main problem with the active handling of structural constraints comes from the fact that the first stage is a costly overhead, as can be seen in the next subsection.

First, let us recall that a *constraint* is a relation over a tuple of variables $\langle v_1, \ldots, v_n \rangle$ which specifies a set $C$ of assignments of $\langle v_1, \ldots, v_n \rangle$, namely the set of the constraint solutions. $C$ is a subset of $D_1 \times \ldots \times D_n$ where $D_i$ $_{(1 \leq i \leq n)}$ is the domain of $v_i$. The sets $D_1, \ldots, D_n$ are specified by the model of the constraint.

**Resolvability preconditions.** The first stage consists in selecting from the current constraint system a subsystem composed of the constraints such that the available partial description of the model allows the solver to decide whether, when considered separately, they are satisfiable or not. In other words, a constraint is *resolvable* if it is possible to determine definitely whether the system composed of this single constraint is satisfiable or not. More formally, a constraint $C$ over $\langle v_1, \ldots, v_n \rangle$ is said to be resolvable if it is possible to associate with each variable $v_i$ $_{(1 \leq i \leq n)}$ a *complete partial domain* $d_i \subset D_i$. $d_i$ is said to be a complete partial domain of $v_i$ if and only if:

1. all $d_i$ members belong to the current *partial* description of the model, and

2. all the values that $v_i$ may have in any of the CLP program solutions are members of $d_i$.

Therefore, the CLP program solutions set remains unchanged if we take $d_i$ instead of $D_i$ as domain of $v_i$.

In order to compute a possible complete partial domain for a variable $v_i$ the constraint solver uses all the constraints which concern $v_i$. Each constraint $C$ on $v_i$ specifies a set:

$$(1) \quad d_i^C = \{a_i \in D_i / \exists a_1, \ldots, a_{i-1}, a_{i+1}, \ldots, a_n \in D_1 \times \ldots \times D_{i-1} \times D_{i+1} \times D_n, \langle a_1, \ldots, a_n \rangle \in C\}$$

If the system can establish that $d_i^C$ is contained in the model current partial description, then it can use it as a complete partial domain of $v_i$

For example, consider the constraint `daughter(Y,x)` which stipulates that `Y` must be a daughter of `x`, where `x` is a node of a tree $\mathcal{T}$ and `Y` is a variable taking its values in the set $\mathcal{N}$ the nodes of $\mathcal{T}$. `daughter(Y,x)` can be used to associate with `Y` the partial domain $\mathtt{d_Y} = \{a \in \mathcal{N} / \mathtt{daughter}(a, \mathtt{x})\}$. If the solver has the information that, in the current partial description, all daughters of `x` have already been attached to `x`, then it can deduce that $\mathtt{d_Y}$ is a complete partial domain for `Y`, and, since `x` is a constant, that `daughter(Y,x)` is resolvable.

**Classic resolution of the resolvable constraints.** In the second stage, the resolvable constraints subsystem is solved by means of standard CLP techniques. The subsystem constraints are indeed different from those in the original system in the sense that their variables are associated with their complete partial domains instead of their original full domains. The constraints resolution has two aims: first, the affectation of a value to the variables that can take only one value and second, the simplification of the constraint subsystem [Cohen90]. For instance, consistency techniques can be used [VanHentenryck89] if all variables have finite domains (e.g. sets of tree nodes). These techniques consists in filtering the variables (complete partial) domains in order to remove the values that do not appear in any of the solutions of the resolvable constraints subsystem (i.e. the subset of the resolvable constraints).

## 2.3 Overhead Cost

Structural constraints can be said to be active only if the second stage of their resolution (affectation of values to the variables and simplification of the constraint subsystem) takes place as soon as possible, that is, as soon as their resolvability precondition holds. The checking of the precondition (first stage) must then be reiterated at every resolution step as long as it does not hold. Actually, it is possible to perform the selection of the resolvable constraints subsystem only when new constraints on the linguistic structure are added. Practically, this strategy may be implemented by means of a typed constraint language (with typed variables and typed domains): the solver is the invoked only when structural variables are affected by new constraints or when modification are made to the model partial description.

The following example gives an idea of the complexity of the computations needed to check resolvability preconditions. Let $u$ and $v$ be variables with finite domains $D_u$ and $D_v$ such that $v$ has a complete partial domain $dom(v)$ of $m$ elements, and that $u$ only appears in a single *binary* constraint $c(u, v)$. The completeness of $dom(u) = \{a \in D_u / \exists x \in dom(v), c(a, x)\}$, the complete partial domain of $u$, is computed according to the following formula:

$$(2) \quad complete(dom(u)) \leftrightarrow complete(dom(v)) \wedge (\forall x \in dom(v), complete(c(x)))$$

where $c(x) = \{a \in D_u / c(a, x)\}$. In the worst case, the right hand side member of (2) is evaluated once when $dom(v)$ becomes complete and once each time a set $c(x)$ becomes complete ($x \in dom(v)$). The number of tests induced by the determination of the completeness of $dom(u)$ is therefore $m(m+5)/2$: $complete(dom(v))$ is tested $m$ times and $complete(c(x))$ is tested $(\sum_{i=0}^{m-1}(i+1)) + m$ times. If the constraint were 3-ary, i.e., $c(u, v, w)$ with $\mid dom(v) \mid = m_v$ and $\mid dom(w) \mid = m_w$, the previous formula becomes:

(3) $complete(dom(u)) \leftrightarrow$

$$complete(dom(v)) \wedge complete(dom(w)) \wedge$$
$$(\forall x, y \in dom(v) \times dom(w), complete(c(x,y)))$$

and the number of tests $m_v m_w (m_v m_w + 5)/2$. The complexity of the computation of the preconditions of a system of binary constraints on $n$ variables having each a complete full domain of $m$ elements is, in the worst case, in $o(nm^2)$. Remember that the preconditions checking are an *overhead* which must be added to the usual amount of computation needed for the standard resolution of the constraints (second stage).

### 2.4 Conclusion

In this section, we have discussed some drawbacks of the active resolution of structural constraints: the need to provide the solver with a description of the model while it is being built, the selection of the resolvable constraints subsystem and the cost of this overhead. Some other questions, such as the definition of the structural constraints languages, their adequacy to NLP and the deductive capability of the structural constraints solvers have not been addressed because of the space limit.

The active handling of the structural constraints has a very heavy cost because of the resolvability preconditions checking and the deductions that the solver must perform. This makes *general* structural constraint resolution much too penalizing to be used in NLP systems. On the other hand, the active handling of linguistic constraints presents numerous advantages when they are always resolvable (i.e they are not be sensible to model underspecification) and when they can come down to constraints on standard CLP domains (boolean, numeral, finite domains). In the remainder of the paper, we present several arguments which support this claim.

## 3 First Example: a Bottom-up Parser

This section presents a toy-parser highlighting some properties of constraint logic programming. It uses a bottom-up strategy which can be implemented in a straightforward and very concise way.

The technique is classic and consists in scanning the input sentence through a window. This window has a varying size and is used for *handle* recognition (this operation is executed without any particular direction). In case of failure, the window's size is increased and scanning starts again. Practically, we can represent the input sentence and the window as lists. The original list (the sentence) is split in three successive sublists, the middle one being the window. The first and the last sublists may be empty. The choice of the first sublist's size determines the origin of the window. The last sublist has no particular role. This is obviously not the better strategy but our intend is present a genuine strategy completely different from usual Prolog implementation of context-free parsers, this way underlining interesting CLP properties such as conciseness.

The implementation is straightforward with list constraints (in particular equality and size constraints pre-defined in Prolog III). The parser chooses different list sizes in order to find handles using an enumeration on the size of the first list (predicate `enum`). The first sublist being fixed, verification of the existence of a handle is performed and the process repeats recursively.

For clarity's sake, the parser input is a category list. The grammar (more precisely the set of phrase-structure rules) is represented by the `ps-rule` clauses. The parser itself is implemented in the `parse` clause. The set of constraints stipulates that the input list is a concatenation of three sublist `a`, `b` and `c`. Each sublist has a given size, respectively `a1`, `b1` and `c1`. The value of `a1` is given by an enumeration predicate, `b1` and `c1` being deduced according to the size constraints. The main process consists only in a recursive call to the `parse` predicate once a handle has been recognized by a `ps-rule` clause.

(4)

```
parse(<Se>,t) -> outl(t);
parse(a.b.c,<r,b>.s) ->
     enum(a1)
     ps-rule(r,b)
     parse(a.r.c,s),
     {a.b.c::l,l>=a1>=0,l>=b1>0,l>=c1>=0,a::a1,b::b1,c::c1,a1+b1+c1=l};

ps-rule(<Se>,<NP,VP>) ->;
ps-rule<NP>,(<Det,Adj,Nm>) ->;
...
```

A possible query can be:

```
parse(<Det,Nm,Vb,Det,Nm,Prep,Nm>,t);
```

```
 {t =  <<NP>, <Det,Nm>, <NP>, <Det,Nm>, <NP>, <Nm>, <PP>, <Prep,NP>, <VP>, <Vb,NP,PP>,
       <Se>, <NP,VP>>}
 {t =  <<NP>, <Det,Nm>, <NP>, <Nm>, <NP>, <Det,Nm>, <PP>, <Prep,NP>, <VP>, <Vb,NP,PP>,
       <Se>, <NP,VP>>}
 {t =  <<NP>, <Det,Nm>, <NP>, <Nm>, <PP>, <Prep,NP>, <NP>, <Det,Nm>, <VP>, <Vb,NP,PP>,
       <Se>, <NP,VP>>}
 ...
```

The results correspond to different derivations (but not different trees) of the input. This example shows that the reduction process can begin anywhere in the sentence.

Obviously, we could refine this core mechanism with the use of an actual control process, for example by introducing specific functions as in DCGs. But it would not alter the fact that this parser works and provides the derivation lists.

# 4 Structural Constraints

We first address the question of identification of basic linguistic constraints remaining as general as possible. In this perspective, we remark that several structural properties need to be verified.

Let us consider that the basic syntactic data structure can be represented as a local tree (i.e. a connected subgraph of a tree). This is a simplified notation, in particular for feature-based theories, but the important point here is the hierarchical relation without any type notion.

So, given a local tree of the form $R(x, y, W)$ where $R$ is the root, $\{x, y\} \cup W$ the daughters with $W$ a possibly empty set of categories. $R$ corresponds to a non-lexical category, $x$ and $y$ are two of its constituents. Here are the most basic well-formedness constraints:

| | | |
|---|---|---|
| (5) | $x \neq y$: all constituents of the same category are different[a]. | |
| (6) | *PL(x,y)*: the sequence $/xy/$ satisfies linear precedence constraints. | |
| (7) | $\{x,y\} \subseteq$ *Legal_Daughters(R)*: the local tree satisfies a dominance schemata[b]. | |
| (8) | *Subcat(R) = Subcat(x)* $\setminus \{y\}$ *(if Proj(x)=S)*: the valency must be satisfied. | |
| (9) | $R \in \{Proj(x), Proj(y)\}$ : $R$ must be a projection of at least one of its constituents. | |

[a] This is a general constraint which must be relaxed in some cases such as conjunctions.
[b] *Legal_Daughters* relies either on right-hand side of PS-rules or on immediate dominance schematas.

These constraints are the most basic and need to be completed with more specific ones during construction of a complete syntactic structure. In the case of feature-based theories, we use more complex mechanisms such as instantiation principles which can be represented with active constraints (see [Blache93]). But, in this section, we focus only on the most general and cross-theoretical properties, the representation of which we describe hereafter. The basic constraints (5) and (9) are the most simple. We will call them *constituent restriction constraints*. They can be represented directly with active constraints. The others are more complex and constitute entire problems we address in the next sections.

## 4.1 Constituent Restriction Constraints

### 4.1.1 Constituent Unicity

As we have seen it, the problem of defining active constraints in NLP comes from the fact that, even if a constraint seems to be clear and simple, the structure of the constrained objects are not always known before the parsing process starts. We do not know for example the exact number of complements governed by a given head beforehand. As proposed in [Guenthner88], a solution consists, in using a canonical syntactic structure to apply constraints *a priori*.

The question now is how can we apply this approach to the *unicity* constraint? This property specifies that all the constituents of a phrase (or a proposition) must be different from each other. The implementation with active constraints is direct using the symbolic constraint `AllDistincts` [1]. This constraint holds if all the elements of a list are distinct from each other. Let us take the case of HPSG. In this theory, the syntactic hierarchy is represented by means of a complex feature called DAUGHTERS which takes as value different signs: HEAD, COMPLEMENT, FILLER, ADJUNCT, etc. The structure (10) constitutes a canonical hierarchy schemata[2].

$$(10)\quad \begin{bmatrix} \text{DTRS} & \begin{bmatrix} \text{HEAD-DTR} : \textit{sign} \\ \text{FILLER-DTR} : \textit{sign} \\ \text{MARKER-DTR} : \textit{sign} \\ \text{COMP-DTRS} : \textit{sign}* \\ \text{CONJ-DTRS} : \textit{sign}* \\ \text{ADJ-DTRS} : \textit{sign}* \end{bmatrix} \end{bmatrix}$$

[1] This constraint is pre-defined in CHIP and can be directly implemented in other languages such as Prolog III.
[2] The value of some of these features such as COMP-DTRS is, in theory, a set of sign; for clarity's sake, we restrict them in this presentation to be a single sign.

The unicity constraint consists then in specifying that the values of the different daughters must all be different. Practically, creating a non-lexical sign resolves to create a new feature structure containing the daughters hierarchy on the basis of the canonical one. The unicity constraint controls this structure and is installed by the creation mechanism. It holds if the daughters cannot unify.

This property can be represented as follows:

(11)

| | |
|---|---|
| *Created structure* : | DTRS[HEAD-DTR(a) ∧ COMP-DTRS(b) ∧ FILLER-DTR(c) ∧ CONJ-DTRS(d) ∧ ADJ-DTRS(e)] |
| *Unicity Constraint* : | `AllDistincts(a,b,c,d,e)` |

Let us notice that this constraint can be restricted to the single category value (represented in HPSG by MAJ feature). In this case, we would need only to replace the argument of the daughters features with the corresponding path. HEAD-DTR(a) would become HEAD-DTR(SYN | LOC | CAT | HEAD | MAJ(a)).

### 4.1.2 Projection Constraint

This property specifies that a non-lexical category must be the *projection* of one of its constituents (i.e. each non-lexical category must have a head). Such a constraint is very important, in particular because of head feature values transmission between these two categories.

The representation of this constraint is straightforward. In the case of unification grammars, this constraint is applied with the unicity one during the creation of a non-lexical feature structure. Within the hierarchical syntactic structure from the previous section, this constraint will enforce the instantiation of the head daughter.

(12)

| | |
|---|---|
| *Created structure* : | DTRS[HEAD-DTR(a) ] |
| *Projection constraint* : | $\mathtt{a} \neq \emptyset$ |

Let us remark that in a canonic feature structure, the empty set value means that the attribute has no value, i.e., that it is absent (the value associated with a feature may be either an atom or a set of attribute/value couples). Incidentally, note that the remark in the previous section about the specification of a particular path for the head daughter applies here also.

## 4.2 Subcategorization

Subcategorization defines a relation between a head and its complements. It describes the different categories (the valency schemata) which can be governed by a head (generally a major category). This general notion is essential to all linguistic theories. However, there exists a lot of variation in its use and implementation: for simple phrase-structure formalisms, subcategorization is partly implemented in the PS rules, and partly as an *a posteriori* verification during lexical insertion; but for lexicalized theories (HPSG or TAG), this notion plays a more active and explicit role where phrase-structure rules are replaced with general schemata.

The representation of subcategorization can vary greatly from one linguistic formalism to another. But for all, we can say that this mechanism consists in constraining the complementation relation by a reduction of the instantiation domain (the set of all categories) to the set of possible complements (for a given head). Notice that we use the notion of category and complement in its most general sense (complementation involves subject-verb and determiner-noun relation as well as modification).

From a computational perspective, this problem concerns the finite domain of possible complements. The subcategorization, which corresponds to a reduction operation, could be represented with symbolic constraints. If so, a parser, for example, (whatever be its strategy, top-down or bottom-up) would only generate values belonging to the appropriate subdomain (by derivation or shift-reduce).

(13) Subcategorization of a transitive verb: `element(x,<NP>)`

(14) Subcategorization of a noun: `element(x,<Det,Adj,PP>)`

where `x` represents the complement of the head. Similarly to the constraints, (13) and (14) are posited on the value of the complements of the head when it is created. Previous examples use the symbolic constraint `element` similar to the one of CHIP.

One problem with this approach (and more generally with subcategorization) lies in the fact that we cannot specify any difference between optional and compulsory complements. This distinction is of course very important from a linguistic perspective: a preposition needs an NP complement within a well-formed PP while a noun can be constructed in an NP without adjective[3]. In a more formal point of view, this issue has interesting consequences: the well-formedness condition of a phrase depends on the realization of its compulsory constituents together with the well-formedness of the realized optional constituents. So, the maximal constituent set ($M$) of immediate constituents of a phrase-level category is the union of the sets of compulsory constituents ($C$) and optional ($O$) constituents. Subcategorization constrains the instantiation on both $C$ and $O$; different valency schemata correspond to subsets of $O$. Therefore, a classic subcategorization schemata is a subset of $M - \{Head\}$.

We present here a method combining the precision of the above-described properties and the efficiency of the finite-domain constraints. To this end, both symbolic and boolean constraints are brought into play. The mechanism consists in associating each category to a boolean value representing its well-formedness.

Phrase-level categories are associated with general schemata whereas subcategorization itself is represented at the lexical level. The schemata contains:

- the definition of the set $M$ of all the immediate constituents,
- the basic well-formedness constraint of a phrase: a phrase is basically well-formed if and only if its compulsory constituents are realized and well-formed.

The set $M$ is given for every PS level category. So, we can use "classic" CLP constraints: $M$ may be used as a partial model for the constraints over the element of $M$ and their mother category.

The realization of compulsory constituents can be represented with a boolean constraint on the well-formedness values indicating that a phrase-structure category is basically well-formed if its compulsory constituents are realized. Subcategorization reduces this set, by specifying the realization of some categories belonging to $M$. The following example is a set of

[3] HPSG describes this distinction with the MOD feature.

Prolog III constraints describing some symbolic or boolean constraints associated with nominal categories:

| | | |
|---|---|---|
| (15) | Constituents : | { M = <N,Det,Adj,PP>} |
| | Compulsory : | {SN ⇒ N} |
| | Subcategorization 1 : | {Det ∧ Adj = true} |
| | Subcategorization 2 : | {Det ∧ PP = true} |

Notice that, by abuse, the same symbol represents both a category and its well-formedness value (a boolean variable).

Similarly, the subcategorization between a predicate and its compulsory arguments can be expressed as: Pred ⇒ Argt where Pred and Argt are boolean representing the realization (and the well-formedness) of the predicate and its argument.

## 5 Value Constraints

The representation of linguistic objects and syntactic relations as feature structures is now very frequent in natural language processing. Several works ([Carpenter92], [Johnson91], [Smolka89]) have described the formal properties of such representations. They define in particular a satisfaction relation between feature structures and constraints on these structures (called descriptions). A CLP approach for NLP relies on the interpretation of linguistic constraints as feature structure descriptions.

We take in this section the case of two specific parsing processes using feature structures: *feature cooccurrence restriction* (hereafter FCR) and *instantiation principles* (noted IP).

The basic parsing mechanism for the linguistic formalisms relying on feature structures comes to specify two kinds of relations: one between feature values and the other between feature structures. The former (such as FCR[4]) are local and concern features belonging to the same structure. The later can be long-distance dependencies and correspond to instantiation principles. In both cases, the mechanism consists in verifying (or instantiating) the value of a particular feature in relation with other feature values.

Let us take some examples.

| | | |
|---|---|---|
| (16) | *FCR* : | PFORM ⊃ ¬ INDEX<br>VFORM ⊃ MAJ[V]<br>+PRD ∨ VFORM ⊃ PAS ∨ PRP |
| | *IP* : | in a headed-structure (i.e. a structure with an instantiated HEAD-DAUGHTER feature) HEAD values of the sign (i.e. the structure) and its head daughter must be token identical. |

We can consider that FCRs are essentially lexical whereas IPs deal with phrase-structure level.

In the issue at hand, much more than for the representation of subcategorization, the problem comes from the complexity of the basic data structures. A feature contains at least two informations: its name (which can be seen as its position in the structure) and its value. While we cannot use directly classic constraints, the representation of feature structure parsing within CLP can have two different kind of solution by means of:

[4] The use of FCR is defined for untyped formalisms. Similar constraints can be used for partial (untyped) implementations of HPSG.

- the definition of high level constraints, or
- an interpretation allowing the representation of complex structures with more simple ones.

In the first case, the solution consists in defining a particular constraint language together with an *ad hoc* solver adapted to feature management. Constraints in such a language would represent dependencies between two sets of features, generally in terms of structure-sharing.

In the second case (chosen here), the main issue is about knowledge representation: we need an interpretation of feature structure behavior within a classic CLP domain (numerical, boolean...). But in both cases, problems come from the representation of relations between different feature structures. Moreover, the syntactic structure representation (i.e. the result of a parsing process) is dynamic and two constrained substructures (i.e. two structures in a dependency relation) can have very different forms depending on the input (the parsed sentence). This underspecification property results in the use of a variable set of features but constraining variables of such structures is difficult, whatever be the type of constraint.

To summarize, the difficulty of representing constraints on feature structures is threefold:

- representation of complex structures,
- choice of the constraint nature, and
- underspecification.

As we have seen it, one of the problems concerning the implementation of attribute-value structures is partial information. Two different solutions exists. The first one consists in converting these structures into fixed-arity lists. It is implemented by most of the systems using feature structure representation and was described by [Guenthner88] and by [Nakazawa88] under the name of "description vectors". This solution was also chosen for the implementation of the HPSG-PL system (see [Popowitch91]). A general study about this question can be found in [Schöter93]. In these systems, the use of fixed-arity structures is justified by efficiency arguments. However, this solution is not really satisfying in particular because using partial information is one of the main advantages of feature structure representations. We propose a second solution relying on features indexation where each feature is addressed by a pointer. Johnson uses an equivalent labeling, but his choice is driven by clarity considerations, not the implementation issue (see [Johnson90]). The example (17) shows the representation of an attribute-value matrix as a conjunction of features. This conjunction is implemented by a list of tuples (the feature-structure list) of the form ⟨`Feature, Index, Value`⟩. The forth argument, `Value`, can represent complex values; when it does, its value is a pointer. Practically, the elements of the feature-structure list are sublist and the position of the sublist into the list corresponds to the index (e.g. the substructure 3 is represented by the third sublist in (18). Note that this representation is quite flat, there is only one level of embedded lists[5].

(17) $$\boxed{1}\begin{bmatrix} \text{CAT } \boxed{2}\begin{bmatrix} \text{HEAD } \boxed{3}\begin{bmatrix} \text{MAJ n} \\ \text{CASE nom} \end{bmatrix} \end{bmatrix} \\ \text{CONTENT } \boxed{4}\begin{bmatrix} \text{INDEX } \boxed{5}\begin{bmatrix} \text{GEN masc} \\ \text{NUM sing} \end{bmatrix} \end{bmatrix} \end{bmatrix}$$

[5] From an implementation point of view, such a representation is very useful because some languages like Prolog III allow a direct access using lists constraints.

(18) [ [⟨ cat, 1, 2 ⟩, ⟨ content, 1, 4 ⟩], [⟨ head, 2, 3 ⟩], [⟨ maj, 3, n ⟩, ⟨ case, 3, nom ⟩], [⟨ index, 4, 5 ⟩], [⟨ gen, 5, masc ⟩, ⟨ num, 5, sing ⟩] ]

Such a representation (*i*) associates explicitly attributes and their values and (*ii*) represents the absolute position of the feature in the general structure. So, we do not need any implicit informations as for fixed-arity representation. Moreover, the use of pointers allows a direct implementation of structure sharing which is very useful in particular for HPSG. Finally, and this is the most important point, we can represent directly partial structures, without any compilation stage and without any loss of performance.

Now, let us explore the consequences of the use of this representation for the verification of feature structure properties (the implementation of the description language). Recently, [Ramsay90] has described a representation associating truth values with atomic features. This information is used to constrain feature values instantiation and to allow the representation of negative values.

We propose to extend this approach in two directions: first, by introducing a third value: "unspecified" (useful in particular for lexical definitions) and second, by generalizing this value specification to all features. More precisely, let $f$ be a function from feature values to $D = \{U, F, T\}$. Let $FS$ be a given feature structure and $t$ be a feature. Then $f_{FS}(t) = T$ when the feature $t$ has a legal value in the $FS$ structure, $f_{FS}(t) = F$ when the value of $t$ cannot be legal in $FS$, and $f_{FS}(t) = U$ otherwise.

When we add this new information to our feature representation, we obtain a new tuple of the form $\langle t, i, v, f_{FS}(t)\rangle$ in which $t$ is the feature name, $i$ its index, $v$ its value and $f_{FS}(t)$ its interpretation.

This new kind of information, encoded within the feature structure, allows a direct expression of constraints on feature structures (i.e. descriptions). Moreover, this approach also allows a direct implementation of negation and disjunction on feature values. In our approach, this is done with boolean constraints. In the next section, we see how descriptions can be represented with such constraints.

(19)

| |
|---|
| `VFORM [PAS]` is represented by the 4-uple ⟨ `VFORM,i,PAS,true` ⟩ |
| `PFORM` ⊃ ¬ `INDEX` is represented by the constraint :<br>⟨ `PFORM,_,_,true` ⟩ ⇒ ⟨ `INDEX,_,_,false` ⟩ |

This last constraint could also have been written:

⟨ `PFORM,_,_,V1` ⟩,⟨ `INDEX,_,_,V2` ⟩ {`V1` ⇒ ¬ `V2`}.

Example (19) shows how a feature structure description (say a constraint) can be represented by a boolean constraint on interpretation values.

To summarize, this representation offers several advantages:

- representation of partial information,
- direct implementation of negation and disjunction, and
- direct implementation of structure sharing.

## 6 Last example: an HPSG Parser

In this section, we present the main characteristics of an HPSG parser implemented following the above-described representations.

## 6.1 Using Prolog III

In this parser, HPSG principles are implemented by means of boolean constraints. The core mechanism is described in rule (20). The main problem of this representation consists in associating the constraints with the structure. This can be done by verifying the presence of the features concerned by a principle into a sign (i.e. the feature structure associated with a category). This mechanism corresponds to simple unification of the corresponding sublist (the list `S`, extracted by the rule `Delta`) and a pattern list. For clarity's sake, the rule (20) only implements the HFP. But the same mechanism can be extended to the other principles (see [Blache93] for example).

(20)
```
IP(a,F,F') →
     Delta(a,F,S)
     Add(S',F,F'),
     S = [ ⟨SYNSEM,a,a1,f1⟩, ⟨LOC,a1,a2,f2⟩, ⟨CAT,a2,a3,f3⟩, ⟨HEAD,a3,a4,f4⟩,
          ⟨SUBJ,a3,a5,f5⟩, ⟨COMPS,a3,a6,f6⟩, ⟨DTRS,a,a8,f8⟩, ⟨HEAD_DTR,a8,a9,f9⟩,
          ⟨SYNSEM,a9,a10,f10⟩, ⟨LOC,a10,a11,f11⟩, ⟨CAT,a11,a12,f12⟩,
          ⟨HEAD,a12,a4,f4⟩ ]
     S' = [⟨HEAD,a3,a4,f4⟩]
     f1&f2&f3&f4&f5&f6&f7&f8&f9 ⇒ f10};
```

This rule computes a "target" feature structure `F'` which satisfies the IP principles from a "source" feature structure `F`. Rule `Delta`[6] extracts the part of the feature structure corresponding to the headed structure (indexed by `a`) and returns the source features status (i.e. realized or forbidden), represented by boolean values. The HFP principle is represented as a boolean constraint on these values. More precisely, the conjunction `f1 & f2 & f3 & f4 & f5 & f6 & f7 & f8 & f9` indicates whether the head feature of the head daughter is instantiated. If so, the target feature (represented in the `S'` list) is added to the general feature structure. The source and target feature values unify (they share the same index `a4`). The same mechanism is applied for the other principles: verification of the source features status by rule `Delta`, instantiation of the target features and completion of the general feature structure (rule `Add`).

This implementation is particularly interesting because of its generality. Each principle has a straightforward interpretation and the integration of new principles simply consists in adding the corresponding constraints.

Finally, notice that boolean constraints do not directly modify the feature structure; they only specify the properties it must have. The changes are performed by user-defined predicates. In this sense, this approach falls in the *Ask & Tell* paradigm [Saraswat90]. The constraint system attached to the rule (20) is composed of constraints of type "ask" whereas "tell" constraints (the ones which modify the model) are implemented as user-defined predicates (e.g. `Add/3`).

## 6.2 Using LIFE

The apparition of new constraint logic programming languages integrating different programming paradigms avoids several drawbacks of more classic constraint languages. Practically, LIFE (see [Aït-Kaci94]) allows the expression of constraints on non fixed-arity structures (the

[6] This operation is implemented with list constraints.

$\psi$-terms); moreover it implements inheritance as a constraint. These properties allows an actual direct interpretation of linguistic formalisms relying on typed feature structures.

(21)

```
:: P: phrase |  (P.synsem.loc.cat.head = X,
                 P.dtrs.headDtr.loc.cat.head = X)
```

(22)

```
:: P: phrase | (  P.synsem.loc.cat.subj = X,
                  P.dtrs.subjDtr = Y,
                  P.dtrs.headDtr.synsem.loc.cat.subj = append(X,Y),
                  P.synsem.loc.cat.comps = U,
                  P.dtrs.compsDtr = V,
                  P.dtrs.headDtr.synsem.loc.cat.comps = append(U,V)).
```

Figures (21) and (22) show the implementation of two HPSG principles (HFP and Valency). This formulation is quite similar to the ones proposed in HPSG-PL or ALE. But there is a deep difference: in LIFE, these constraints are actual active constraints applied *a priori*, so avoiding a generate-and-test method. In the other approaches, these constraints are compiled into classic Prolog rules and then become passive constraints.

## 7 Conclusion

Constraint logic programming can be a very efficient tool for natural language processing in several aspects. We have underlined in particular the concision and control properties of the paradigm. But the most interesting property lies in the straightforward interpretation of linguistic theories using actual active constraints. We have shown that such an interpretation can be done without any compilation stage: neither the structures nor the constraints need an *ad hoc* mechanism for their implementation/representation.

Representing the parsing problem as a CSP comes to proposing an entirely new conception of this problem. Indeed, parsing a sentence turns out to be the verification of syntactic structures coherence, but in a particular way: lexical insertion instantiates lexical structures and constraint propagation verifies phrase-level structures.

This paper has focused on parsing problems, but the considered approach provides very general mechanisms and allows for the development of reusable systems. Practically, we have experimented the integration of a prosodic level to a CLP-based parser. The integration of new data and principles has been done directly, without any consequence on the parser architecture.